\documentclass{article}

\usepackage{arxiv}

\usepackage[utf8]{inputenc} 
\usepackage[T1]{fontenc}    
\usepackage{hyperref}       
\usepackage{url}            
\usepackage{booktabs}       
\usepackage{amsfonts}       
\usepackage{nicefrac}       
\usepackage{microtype}      
\usepackage{lipsum}
\usepackage{graphicx}

\usepackage{times}  
\usepackage{helvet}  
\usepackage{courier}  
\usepackage{natbib}  
\usepackage{caption} 

\usepackage{amsmath}
\usepackage{amssymb}
\usepackage{mathtools}
\usepackage{amsthm}
\usepackage{multirow}
\usepackage{diagbox}
\usepackage{booktabs} 

\usepackage[misc]{ifsym}

\graphicspath{ {./images/} }

\title{Exploring the Global-to-Local Attention Scheme in Graph Transformers: An Empirical Study}

\author{
 Zhengwei Wang \\
  School of Computer Science and Engineering\\
  Northeastern University\\
  Shenyang, China \\
  \texttt{2372072@stu.neu.edu.cn} \\
   \And
 Gang Wu \Letter \\ 
  School of Computer Science and Engineering\\
  Northeastern University\\
  Shenyang, China \\
  \texttt{wugang@mail.neu.edu.cn} \\
}

\begin{document}
\maketitle
\begin{abstract}
Graph Transformers (GTs) show considerable potential in graph representation learning. 
The architecture of GTs typically integrates Graph Neural Networks (GNNs) with global attention mechanisms either in parallel or as a precursor to attention mechanisms, yielding a local-and-global or local-to-global attention scheme. 
However, as the global attention mechanism primarily captures long-range dependencies between nodes, these integration schemes may suffer from information loss, where the local neighborhood information learned by GNN could be diluted by the attention mechanism. 
Therefore, we propose G2LFormer, featuring a novel global-to-local attention scheme where the shallow network layers use attention mechanisms to capture global information, while the deeper layers employ GNN modules to learn local structural information, thereby preventing nodes from ignoring their immediate neighbors. 
An effective cross-layer information fusion strategy is introduced to allow local layers to retain beneficial information from global layers and alleviate information loss, with acceptable trade-offs in scalability. To validate the feasibility of the global-to-local attention scheme, we compare G2LFormer with state-of-the-art linear GTs and GNNs on node-level and graph-level tasks. 
The results indicate that G2LFormer exhibits excellent performance while keeping linear complexity.
\end{abstract}


\section{Introduction}

Graph Neural Networks (GNNs), while serving as the dominant approach for graph representation learning, inherently suffer from \emph{over-smoothing} and \emph{over-squashing} issues \cite{alon2021on}. 
Consequently, Graph Transformers (GTs) have recently gained research attention \cite{Chen22a}, as their global attention mechanisms \cite{vaswani2017attention} can mitigate these issues by enabling direct long-range dependency modeling without relying on stacked local aggregation. 
Nevertheless, recent analyses reveal that GTs may overemphasize distant neighbors, a limitation termed the \emph{over-globalizing} problem \cite{pmlr-v235-xing24b}.


\begin{figure}[!hbpt]
	\centerline{\includegraphics[width=0.45\textwidth]{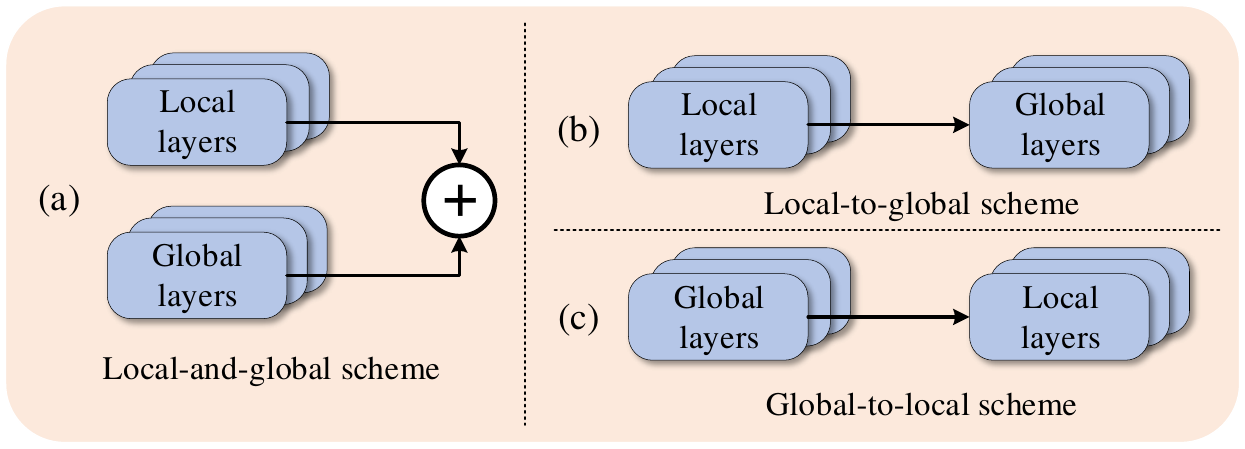}}
	\caption{Categories of attention schemes in which local layers stand for GNN and global layers stand for the attention mechanism. 
    (a) Local layers and global layers learn node representations in parallel and aggregate their outputs in a specific way;
    (b) Local-to-global: Local layers precede global layers, with the attention mechanism learning the final node representations;
    (c) Global-to-local: Global layers precede local layers, with GNN learning the final node representations. 
    Notably, while (a)-(b) are established approaches, scheme (c) remains underexplored.}
	\label{schemes}
\end{figure}

Integrating GNNs with the attention mechanism in GTs has become a prevalent approach, where GNNs model local information while the attention mechanism captures global dependencies  \cite{ma2024polyformer,10479175}. 
There are two primary schemes for integrating GNNs and GTs as shown in Figure \ref{schemes} (a) and (b): the \emph{local-and-global} scheme where local layers and global layers output in parallel, and the \emph{local-to-global} scheme where local layers precede global layers. 
Although both schemes demonstrate comparable performance with no consensus on superiority \cite{min2022transformergraphsoverviewarchitecture}, and have even developed linear GTs \cite{wu2023sgformer, deng2024polynormer} with effective efficiency improvements, they still exhibit inherent limitations.
First, in the local-and-global scheme, the integration method (summation or concatenation) is believed to be oversimplified and fails to facilitate effective information exchange \cite{ijcai2023p501}. 
Second, the local-to-global scheme is considered inferior due to inherent GNN limitations that can cause information loss and further propagation of noisy/erroneous features to downstream global layers \cite{rampasek2022GPS}.
Third, the over-globalizing tendency in GTs introduces structural imbalance: global attention in deeper layers exhibits a preference for distant node interactions, with comparatively insufficient utilization of local patterns from shallow layers, thereby compromising local feature extraction in the representation learning.

The above limitations of existing GTs motivate us to consider an innovative  scheme \emph{global-to-local} (Figure \ref{schemes} (c)), where global attention layers in the shallow parts of networks capture global information while local layers in the deeper parts of networks extract local patterns.
In this paper, we propose an expressive yet efficient GT model, G2LFormer (Global-to-Local Transformer), exploring the global-to-local attention scheme.
To prevent information loss between GTs and GNNs (as in the local-to-global scheme), a cross-layer information fusion strategy is developed to extract information from different layers, facilitating effective learning of both global and local patterns.
We conducted experiments on node-level and graph-level tasks and compared G2LFormer with state-of-the-art GTs and GNNs. 

Our contributions can be summarized as follows.

\begin{itemize}	
	\item 
    To our knowledge, this is the first work investigating a global-to-local feature learning scheme. 
    While adopting the standard practice of using attention for global information and GNNs for local structure, our key innovation lies in the top-down propagation of global-layer information to local layers. 
    This enables better capture of long-range dependencies and refined local aggregation simultaneously.
	
	\item 
    To prevent information loss between network layers, we employ a novel cross-layer information fusion strategy that dynamically balances global and local information, which adaptively reallocates node weights, and effectively addresses both the over-smoothing problem in local layers and the over-globalization issue in attention layers. 
    Remarkably, the entire fusion process does not increase model complexity, demonstrating that enhanced cross-layer interaction need not compromise computational efficiency.
	
	\item 
    We propose G2LFormer as an implementation of the global-to-local attention scheme and employ an existing attention mechanism (derived from SGFormer \cite{wu2023sgformer}) and a GNN (Cluster-GCN \cite{clustergcn} or GatedGCN \cite{gatedgcn}) as the backbones. 
    Experiments show that G2LFormer achieves state-of-the-art performance on both node-level tasks and graph-level tasks. 
    We also prove the linear complexity of G2LFormer theoretically and experimentally.
\end{itemize}

\section{Related Work}
\textbf{Graph Neural Networks.} 
Message passing neural network (MPNN) is a general paradigm of GNNs, which utilizes message passing to propagate and aggregate information from local structures, enabling it to achieve excellent performance in graph representation learning \cite{pmlr-v70-gilmer17a}. 
Numerous GNN advancements are derived from classic GNNs such as graph convolutional network (GCN) \cite{kipf2017semisupervisedclassificationgraphconvolutional} and Graph Attention Network (GAT) \cite{velickovic2018graph}, which are popular baselines for graph representation learning. 
Classic GNNs are the fundamental models for adapting graphs and have already undergone extensive algorithmic optimizations \cite{NEURIPS2022_1385753b}. 
Currently, adapting to large-scale graphs is a challenge for graph mining models. 
For large-scale graph processing, GNN-based approaches are mainly aimed at optimizing training strategies \cite{10.1145/3580305.3599565, 10.1007/978-3-031-26390-3_22, clustergcn}.
Cluster-GCN is such an improvement on GCN, incorporating graph partitioning and local convolution, tailored for large-scale graph adaptation \cite{clustergcn}. 
To address the drawbacks of over-smoothing and over-squeezing of GNNs, GatedGCN \cite{gatedgcn} augments standard graph convolutions with gating mechanisms to dynamically regulate message passing between nodes. 
Furthermore, to effectively transmit information within the network layers, a cross-layer information fusion strategy is proposed in \cite{wang2024nodescreatedequalnodespecific}, which is employed in this work to selectively filter and propagate beneficial information between global layers and successive local layers.

\textbf{Graph Transformers.} 
Early GTs suffered from limited scalability due to their quadratic complexity and exhibited relatively rudimentary expressivity \cite{shehzad2024graphtransformerssurvey}. 
The high memory requirements of complex datasets highlight the importance of scalability, prompting a growing number of GTs to conduct experimental evaluations on complex datasets \cite{kong2023goat, zhang2024torchgtholisticlargescalegraph}. 
To balance scalability and expressivity, algorithm-level optimizations are prevalent.
Some GTs tend to exhibit linear complexity while maintaining performance, and run efficiently on large-scale graphs \cite{ma2024polyformer, yang2022hypformer}. 
However, prioritizing running speed alone is not the optimal solution, as overly concise propagation formulas are not conducive to the model learning complex node representations \cite{Li2024Rethinking}. 
Moreover, few researchers have yet discussed the categories of attention schemes at the model-level. 
Despite being regarded as having limitations, the local-to-global scheme and the local-and-global scheme remain two primary attention schemes \cite{ijcai2023p501}. 
The local-to-global attention scheme is the most commonly employed architecture in early GTs, represented by GraphTrans \cite{NEURIPS2021_6e67691b}. The local-and-global attention scheme subsequently garners increased scrutiny as the sequential transmission between local layers and global layers is believed to cause information loss, with GraphGPS being a notable representative of this scheme \cite{rampasek2022GPS}. 
Nevertheless, the approach of parallel output between local layers and global layers within the local-and-global attention scheme is criticized for being simplistic \cite{ijcai2023p501}. 
At present, the representative GTs for the local-and-global attention scheme and the local-to-global attention scheme are SGFormer \cite{wu2023sgformer} and Polynormer \cite{deng2024polynormer}, respectively. SGFormer proposes a linear attention mechanism that obviates approximation processing, demonstrating high representational capacity with a single attention network layer. 
Polynormer employs polynomial functions in Transformer's formula, streamlining the attention mechanism while exhibiting superior performance. 
The aforementioned two models make a comprehensive trade-off between expressivity and scalability. The attention mechanism of SGFormer has extremely low overhead and is employed as our backbone for global layers.

\section{Preliminaries}
Assume that $\mathcal{V}$ and $\mathcal{E}$ are the node set and the edge set of graph $\mathcal{G}$ respectively.
A graph can then be defined as $\mathcal{G} = (\mathcal{V},\mathcal{E})$. 
Here, $X \in \mathbb{R}^{N \times d}$ denotes the node feature matrix of $\mathcal{V}$, where $N=|\mathcal{V}|$ is the number of nodes and $d$ is the dimension of features. 
$A$ is the adjacency matrix of $\mathcal{G}$.

\textbf{GNN.} 
Assuming that $h^l_v$ is the representation of a node $v\in\mathcal{V}$ in the $l$-th layer of a GNN, and $h^0_v$ is the initial node feature $X_v$. 
$\mathrm{AGG}^l$ is the aggregate function of the $l$-th layer that aggregates the representations of the node's neighbors, and $\mathrm{UPDATE}^l$ is the function that updates the node representations in the $l$-th layer. 
Then the representation $h^{l}_v$ is given by Equation \ref{EQU:node-representation}, where $\mathcal{R}\left(v\right)$ represents the neighbor set of node $v$.

\begin{equation}\label{EQU:node-representation}
h^{l}_v = \mathrm{UPDATE}^l \left(h^{l-1}_v, \mathrm{AGG}^l\left(\left\{h_u^{l-1}\middle| u\in\mathcal{R}\left(v\right)\right\}\right)\right)
\end{equation}

\textbf{Attention mechanism.} 
GTs build upon the Transformer architecture \cite{vaswani2017attention}, whose core operation is computing self-attention. 
The standard self-attention mechanism consists of three learnable matrices, i.e. the query (Q), key (K) and value (V) matrices. 
The three matrices are calculated based on node features $X$ and specific weights as in Equation \ref{EQU:QKV}, where $W_Q$, $W_K$, $W_V$ $\in\mathbb{R}^{d \times d}$ are the weight matrices corresponding to $Q$, $K$, and $V$.

\begin{equation}\label{EQU:QKV}
	Q=XW_Q, K=XW_K, V=XW_V
\end{equation}

For simplicity of illustration, the dimensions of $Q$, $K$, and $V$ are assumed to be the same as $d$. Then the equation of self-attention score without bias terms $Attn$ is as follows:

\begin{equation}
	Attn=\mathrm{softmax}\left(\frac{QK^T}{\sqrt d}\right)V
\end{equation}

\section{Global-to-local attention scheme}

\subsection{Motivation} 
Both local-and-global and local-to-global places global layers near the deep parts of networks, resulting in global-layer-dominated learning of final node representations.
According to recent evaluations on GTs, the over-globalizing problem prevents nodes from sharing the same label with their proximate neighbors \cite{pmlr-v235-xing24b}, which means that global layers underutilize local information.
Thus, a proposal is put forward: to adopt a global-to-local attention scheme, i.e., positioning global layers in the shallow parts while placing local layers in the deep part, thereby alleviating the deficiencies of the attention mechanism.

For the global-to-local attention scheme, global information serves as important prior knowledge, which is beneficial for distinguishing different nodes with similar local structures, thereby assisting local layers. 
Meanwhile, placing local layers in the deep parts of networks alleviates the over-globalizing problem in GTs, as GNNs enhance the learning of local structures. 
Nevertheless, the global-to-local scheme faces the same challenge as the local-to-global scheme, where local layers may lose information transmitted from global layers in the shallow parts of the network architecture. 
It is crucial to incorporate a cross-layer information fusion strategy that spans the entire network to maintain appropriate node representations. 
Moreover, to be applicable to complex datasets, the global-to-local scheme should incorporate scalable GT and GNN in its design.

\subsection{Architecture} 
\begin{figure}
	\centerline{\includegraphics[width=0.4\textwidth]{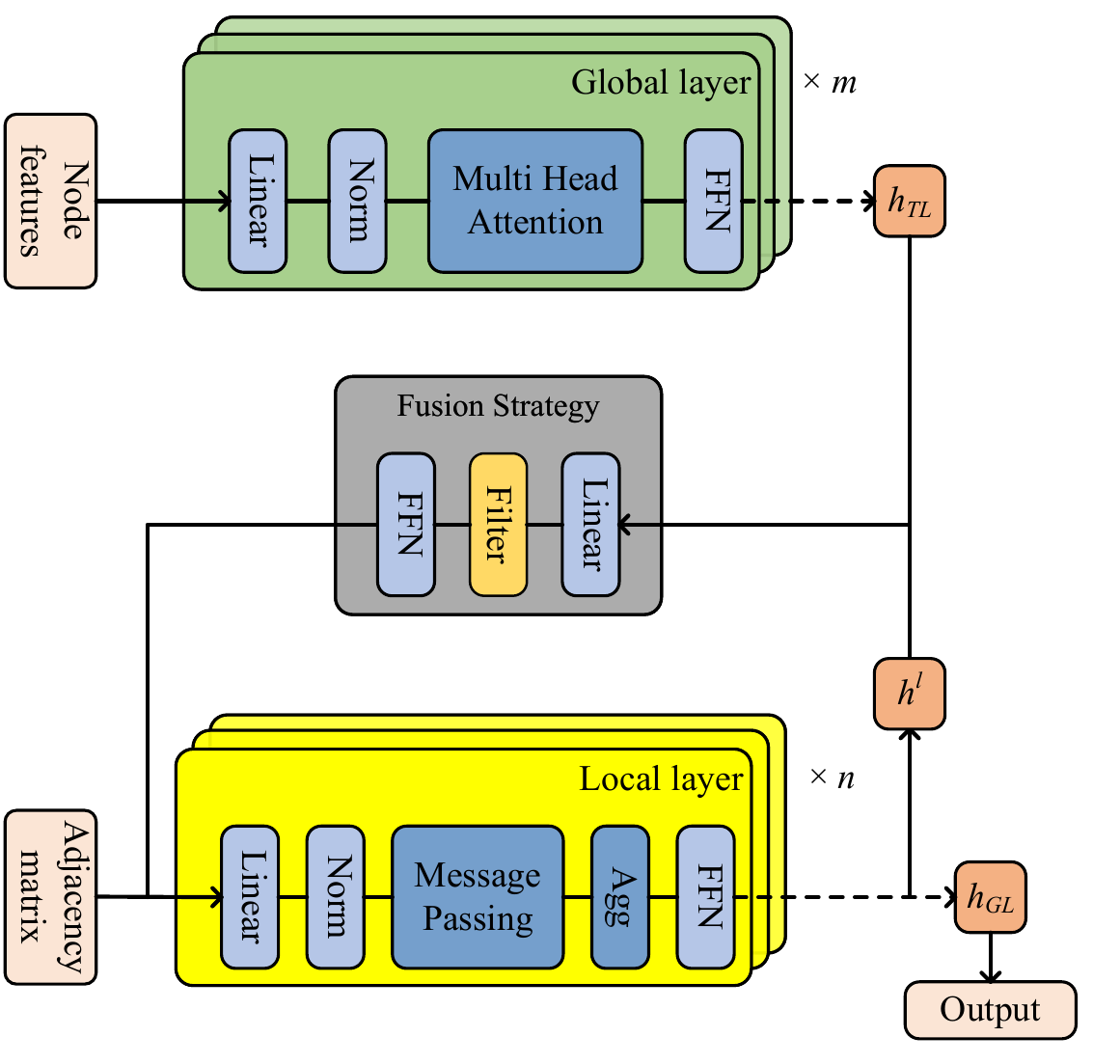}}
	\caption{
    The general framework of global-to-local attention scheme. 
    ``Filter'' is equivalent to the operation of preserving critical node information in cross-layer information fusion strategy. 
    Both local and global layers are modular components that admit substitution with other backbone models.
    }
	\label{framework}
\end{figure}

A general architecture of the global-to-local attention scheme is illustrated in Figure \ref{framework}, which uses $m$ global layers to learn node representations first and then transmits global information to $n$ local layers to learn local structures. 
The node representation learned by global layers is denoted as $h_{TL}$, and the final node representation learned by local layers is denoted as $h_{GL}$. 
Both global layers and local layers contain mappings such as linear projection and feed-forward neural network (FFN). 
In this processing flow, the cross-layer information fusion strategy preserves effective information, acting as an information filter to prevent GNNs from losing information due to over-smoothing and over-squashing. 
Note that the global-to-local attention scheme can also be implemented through interleaved stacking of global and local layers, with cross-layer information fusion strategies connecting the stacked modules. 
 
As a concrete implementation of the global-to-local scheme, we propose G2LFormer, a versatile model adaptable to diverse downstream tasks that employs Cluster-GCN \cite{clustergcn} or GatedGCN \cite{gatedgcn} as the backbone of local layers and the global attention component of SGFormer \cite{wu2023sgformer} as the backbone of global layers. 

\subsection{G2LFormer}
\subsubsection{Global layers}
G2LFormer adopts SGFormer's simplified attention mechanism, which does not require approximation and has been theoretically proven to capture long-range dependencies with just one layer and single-head.
In G2LFormer, the number of layers of the attention mechanism is fixed at 1, i.e., $m=1$. 
G2LFormer obtains $Q$, $K$, and $V$ by applying linear transformations to the node feature matrix $X$.
Let $f_Q$, $f_K$, and $f_V$ denote the linear transformation operations (each a linear feed-forward neural layer in the implementation) for generating $Q$, $K$, and $V$, respectively, then the linear attention function of G2LFormer is computed as follows:

\begin{equation}\label{equ-QKV}
	Q=f_Q(X), K=f_K(X), V=f_V(X)
\end{equation}
\begin{equation}\label{equ-normalize}
	\tilde{Q}=\frac{Q}{\|Q\|_F}, \tilde{K} = \frac{K}{\|K\|_F}
\end{equation}
\begin{equation}\label{equ-denominator}
	\mathcal{D} = \text{diag}^{-1}\left({\mathrm{I} + \frac{1}{N} \tilde{Q} (\tilde{K}^\top \mathbf{1})}\right)
\end{equation}
\begin{equation}\label{equ-update}
	h_{TL} = \mathrm{FFN} \left(\mathcal{D} \cdot \left(V + \frac{1}{N} \tilde{Q} (\tilde{K}^\top V)\right) \right)
\end{equation}

First, Equation \ref{equ-QKV} computes $Q$, $K$, and $V$ with corresponding linear transformation functions.
Then, in Equation \ref{equ-normalize}, $Q$ and $K$ are normalized to $\tilde{Q}$ and $\tilde{K}$ using Frobenius norm $\| \cdot \|_F$. 
The result of Equation \ref{equ-denominator} is used as the normalization factor for global attention, where $\mathbf{1} \in \mathbb{R}^{D \times 1}$ denotes a $D$-dimensional column vector of ones, and $\mathrm{diag}()$ replaces an $N$-dimensional column vector with an $N \times N$ diagonal matrix.
As shown in Equation \ref{equ-update}, the normalization factor $\mathcal{D}$ is applied to update the embedding of $V$ with global attention as $V + \frac{1}{N} \tilde{Q} (\tilde{K}^\top V)$ through multiplication. 
The global layers' output representation $h_{TL}$ is then computed using a feed-forward neural network (FFN), which includes activation functions and mapping operations.
Equations \ref{equ-denominator} and \ref{equ-update} decompose the computation of all-pair similarities in the traditional attention design into two successive steps while preserving the equivalence of expressivity, thereby achieving a reduction in the computation complexity from $O(N^2)$ to $O(N)$.

\subsubsection{Cross-layer information fusion strategy}
To preserve valuable information from $h_{TL}$ in local layers, the cross-layer information fusion strategy is introduced before local layers. 
Specifically, G2LFormer employs NOSAF (Node-Specific Layer Aggregation and Filtration) \cite{NEURIPS2022_092359ce, wang2024nodescreatedequalnodespecific} to ensure effective information flow to the next layer, thereby preventing over-smoothing in local layers, while simultaneously reallocating node weights to preserve critical information of global layers and mitigate over-globalization.
NOSAF redesigns node aggregation, including the refinement of node weights, which is applicable to improve weight allocation in attention mechanisms. 
The process of reallocating node weight is as follows:

\begin{equation}
	\beta ^l= 
	\begin{cases}
		h_{TL}W_h^l \parallel \mathbf{0}, & l=1 \\
		\eta^lW_{\eta}^l \parallel h^{l}W_h^l, & 1<l<n+1
	\end{cases}
    \label{equ-information-importance}
\end{equation}
\begin{equation}
	\gamma^{l}=\operatorname{sigmoid}\left(\operatorname{LeakyRelu}\left(\beta^{l} W_{1}^{l}+b_{1}^{l}\right) W_{2}^{l}+b_{2}^{l}\right)
    \label{equ-node-importance}
\end{equation}

In Equation \ref{equ-information-importance}, $\beta^l$ is a matrix that stores the aggregated information of nodes up to the $l$-th layer, and $\mathbf{0} \in \mathbb{R}^{N \times d'}$ is a zero matrix. 
Here, $\eta^{l}$ is a temporary variable used to retain information before the $l$-th layer and is initially assigned the value of $h_{TL}$. 
The operator $\parallel$ represents the concatenation, and $W_h^l$ and $W_{\eta}^l \in \mathbb{R}^{d \times d'}$ are the learnable weight matrices associated with $h^l$ and $\eta^l$, respectively. 
With $\beta^l$, we can then generate the node importance matrix $\gamma^{l}$ according to Equation \ref{equ-node-importance}.
$W_{1}^{l} \in \mathbb{R}^{2d' \times d''}$ and $W_{2}^{l} \in \mathbb{R}^{d'' \times 1}$ are the weight matrices of the first linear transformation and the second linear transformation, respectively, and $b_{1}^{l} \in \mathbb{R}^{N \times d''}$ and $b_{2}^{l} \in \mathbb{R}^{N \times 1}$ are the bias terms. 
Note that $d'$ and $d''$ are customizable dimensions. 
For node $v$, $\beta _v^l \in \mathbb{R}^{2d'}$ encodes node $v$'s aggregated information, and generates its importance score $\gamma_{v}^{l} \in (0,1)$. 

Subsequently, $\gamma^{l}$ updates the node contributions in deeper network layers, and the operation is denoted as $\mathcal{F}_{f}$ (i.e., the filter), which is described by Equation \ref{equ-filter}. 
Equation \ref{equ-filter} implies that the information in node representations is filtered by the cross-layer information fusion strategy, where $\mathcal{B}\left(\gamma^{l}\right)$ is a broadcast function that maps $\gamma^{l}$ from $\mathbb{R}^{N}$ to $\mathbb{R}^{N \times d}$, and $\circ$ represents the Hadamard product. 
The operation iteratively updates each network layer, extracts features from previous network layers, and stores the information in $\eta^{l+1}$ as shown in Equation \ref{equ-11}. 
\begin{equation}
	\mathcal{F}_{f}\left(h^{l}, \gamma^{l}\right)=h^{l} \circ \mathcal{B}\left(\gamma^{l}\right)
    \label{equ-filter}
\end{equation}
\begin{equation}\label{equ-11}
	\eta^{l+1}=\eta^{l} + \mathcal{F}_{f}\left(h^{l}, \gamma^{l}\right)
\end{equation}

\begin{figure}[!hbpt]
	\centerline{\includegraphics[width=0.45\textwidth]{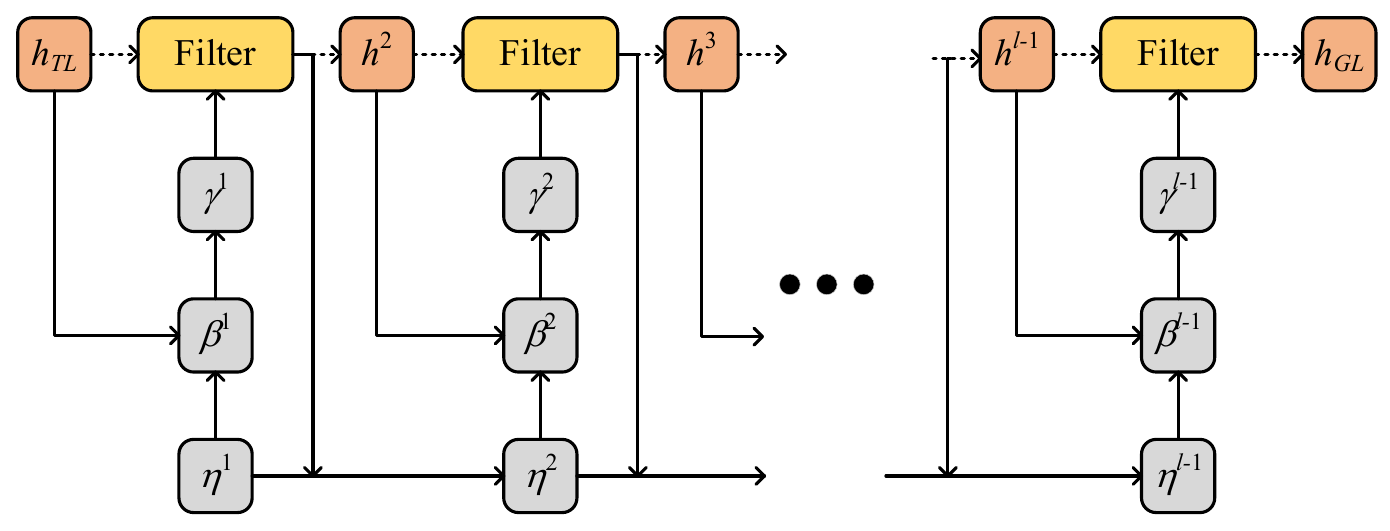}}
	\caption{The  process of the cross-layer information fusion strategy adopted by G2LFormer. The ``Filter'' corresponds to $\mathcal{F}_{f}$, which is identical to the ``Filter'' depicted in Figure \ref{framework}.}
	\label{strategy}
\end{figure}

In summary, the cross-layer information fusion strategy concatenates information from the current layer with previous layers (Equation \ref{equ-information-importance}), then computes the node weights (Equation \ref{equ-node-importance}), and finally updates the node representations accordingly to effectuate filtering (Equations \ref{equ-filter} and \ref{equ-11}). 
The process of the cross-layer information fusion strategy in G2LFormer is illustrated in Figure \ref{strategy}.

\subsubsection{Local layers}
G2LFormer employs two distinct GNN backbones to enhance local structural learning for diverse tasks. 
For node classification tasks, we use Cluster-GCN as backbone to improve scalability by partitioning the graph into subgraphs, performing independent graph convolutions on each subgraph, and merging the results while preserving the original update and aggregation functions. 
For graph-level tasks (e.g., graph classification, graph regression, and inductive node classification) we adopt GatedGCN as it extends standard GCN by introducing effective edge-wise gating mechanisms that enable dynamic control over message passing between nodes. 
The gating function takes node features and optional edge features as input to generate a gating value, which modulates the messages originating from the nodes. 

For clarity, we uniformly define the computation of the $l$-th local layer representation $h_{GL}^{l}$ as in Equation \ref{EQU:local-layers}, where $\tilde{A}$ represents the normalized adjacency matrix and $W^{l-1}$ is the weight matrix in the $l-1$-th layer. 
Here, $h_{GL}^{l}$ requires further processing via the cross-layer information fusion strategy, denoted as $\mathcal{F}_{f}\left(h_{GL}^{l}, \gamma^{l}\right)$, as described in Equation \ref{equ-information-importance} to \ref{equ-11}. 

\begin{equation}\label{EQU:local-layers}
	h_{GL}^{l} = \mathrm{FFN}\left( \tilde{A} h_{GL}^{l-1} W^{l-1} \right)
\end{equation}

Therefore, the complete expression of $h^l$ in G2LFormer is given by Equation \ref{equ-hl}.
The formulation indicates that in G2LFormer, $h_{GL}$ continuously retains the information from previous layers and learns local structures after capturing global attention from $h_{TL}$. 

\begin{equation}
h^{l}  = \begin{cases}

h_{TL}, & l=1 \\
\mathcal{F}_{f}\left(h_{GL}^{l}, \gamma^{l}\right), & 1<l<n+1 \\
h_{GL}, & l=n+1

\end{cases}
\label{equ-hl}
\end{equation}


\subsection{Complexity analysis} 
The computational complexity primarily stems from the instances of global attention and GNN employed. 
Specifically, the complexity is $O(N + 2\vert \mathcal{E}\vert) \approx O(N + \vert \mathcal{E}\vert)$, where GNN contributes $O(\vert \mathcal{E}\vert)$ and global attention contributes $O(N + \vert \mathcal{E}\vert)$. 
Although the cross-layer information fusion strategy involves re-learning aggregated information, we prove in the Supplementary Material that its complexity (Equation \ref{equ-node-importance}) is $O(Nd'd'')$, which further reduces to $O(N)$ when $d', d'' \ll N$.
The sparsity of graphs in the benchmark datasets ensures  $\vert \mathcal{E}\vert$ proportional to $N$ and $\vert \mathcal{E}\vert \ll N^2$, simplifying the global attention matrix calculation. 
Consequently, compared to existing attention schemes, the global-to-local attention scheme introduces no additional computational complexity. 
Thus, G2LFormer achieves computational efficiency equivalent to a linear GT, avoiding node representation information loss at an acceptable additional cost.



\section{Experiments}

We empirically validate the implementation of the global-to-local scheme through comprehensive experiments on two fundamental downstream tasks: node-level task (i.e., node classification) and graph-level task (i.e., graph classification, graph regression, and inductive node classification). 
Benchmark comparisons include state-of-the-art graph neural networks (GNNs) and graph transformers (GTs).

\subsection{Datasets}
Our experiments employ standard benchmark datasets spanning both node-level and graph-level learning tasks. 

For node classification tasks, we select large-scale graphs commonly used in the literature. 
The social network dataset \textbf{Pokec} \cite{Leskovec_Krevl_2014} is chosen for its characteristics as a large-scale heterogeneous graph. 
From the Open Graph Benchmark (OGB) \cite{hu2020ogb}, we consider three representative graphs: \textbf{ogbn-arxiv}, \textbf{ogbn-proteins}, and \textbf{ogbn-products}, which cover different domains and scales.
\textbf{ogbn-products}, also known as \textbf{amazon2m}, is an Amazon product co-purchasing network. 
\textbf{ogbn-proteins} is a protein interaction network. 
\textbf{ogbn-arxiv}, a citation network comprising 0.16 million nodes and 1.1 million edges, is the only dataset in this study that permits full-batch training.
Specifically, for \textbf{ogbn-arxiv}, we adopt the TAPE framework \cite{he2024harnessingexplanationsllmtolminterpreter} and use the node features it provides.
The reason is that TAPE employs a large language model to enrich the node features of the text attribute graphs, which improves the training speed by fine-tuning the original texts through the language model, and achieves state-of-the-art performance in \textbf{ogbn-arxiv}. 

For graph-level tasks, we use the Long Range Graph Benchmark (LRGB) (Dwivedi et al. 2022), a collection of datasets characterized by large graph sizes, longer effective diameters, and the need to capture long-range dependencies, which present a significant challenge for existing models that primarily focus on local neighborhood aggregation. 
The selected datasets include \textbf{Peptides-struct}, \textbf{Peptides-func}, \textbf{PascalVOC-SP}, and \textbf{COCO-SP}. 
We also evaluate on OGB's \textbf{ogbg-molhiv}, a benchmark for molecular property prediction in drug discovery.

\subsection{Baselines}
Since both expressivity and scalability are our primary criteria, we select baseline models with proven effectiveness on large-scale graphs.
We exclude early GTs and some advanced GNNs from our consideration, as they fail to strike a balance between expressivity and scalability, particularly exhibiting poor scalability on complex datasets.

For node classification tasks, we compare our method with classic GNN baselines (GCN and GAT \cite{velickovic2018graph}), which are widely adopted due to their moderate complexity and satisfactory expressivity. 
Furthermore, SIGN \cite{frasca2020signscalableinceptiongraph} and SGC \cite{pmlr-v97-wu19e} are considered, as they are GNNs that focus on improving scalability. 
Cluster-GCN \cite{clustergcn} is not included in our results, as it shows negligible accuracy difference from baseline GCN.
Among various GT variants, DIFFormer \cite{wu2023difformer}, SGFormer \cite{wu2023sgformer}, and Polynormer \cite{deng2024polynormer} emerge as notable representatives. 
These linear GTs offer not only straightforward implementation but also enhanced expressivity and scalability.

For graph-level tasks, we adopt three classic GNN baselines: GCN, GIN \cite{xu2019powerfulgraphneuralnetworks}, and GatedGCN.
As competing graph transformer baselines, we select GraphGPS \cite{rampasek2022GPS}, Exphormer \cite{shirzad2023exphormer}, and GRIT \cite{ma2023GraphInductiveBiases} based on their demonstrated superiority in graph-level tasks.
We also include GECO \cite{sancak2024scalableeffectivealternativegraph}, an enhanced model that replaces the attention mechanism, representing a novel variant of GTs. 
To capture long-range dependencies, we additionally evaluate state space models through two representative implementations: GPS+Mamba \cite{Behrouz2024KDD-GMN} and GMN \cite{Behrouz2024KDD-GMN}.

For all baselines, we use the experimental results reported in their original papers if the dataset settings (e.g., data splitting, batch size, node features) are identical to ours. 
Otherwise, we re-implement those baselines and obtain the optimal results under the same dataset settings.

\subsection{Setup}
The data splitting for each dataset is based on existing practices. 
Specifically, \textbf{ogbn-arxiv}, \textbf{ogbn-proteins}, and \textbf{ogbg-molhiv} are split according to OGB standard settings. 
All datasets from LRGB are split following the standard protocol.
\textbf{ogbn-arxiv} is chronologically partitioned, with papers published until 2017 allocated for training, those published in 2018 for validation, and papers from 2019 reserved for testing.
\textbf{ogbn-proteins} is split based on the species of proteins. 
\textbf{ogbn-products} and \textbf{pokec} datasets are processed following recent work \cite{wu2022nodeformer, wu2023difformer, wu2023sgformer}. 
The data splitting ratio of \textbf{ogbn-products} is 50\% / 25\% / 25\% (train / val / test), while  \textbf{pokec} follows a  10\% / 10\% / 80\% split. 

For the training methods, full-batch training is used for \textbf{ogbn-arxiv}, while random mini-batch training is used for other datasets. 
For node classification tasks, the number of epochs for all datasets is set to 1000 unless overfitting occurs. 
For graph-level tasks, the number of epochs is set between 100 and 300.

We employ the following evaluation metrics: 
ROC-AUC for \textbf{ogbn-proteins} and \textbf{ogbg-molhiv};
Accuracy for other node classification tasks;
Average precision for \textbf{Peptides-func};
Mean absolute error (MAE) for \textbf{Peptides-struct};
F1-score for \textbf{PascalVOC-SP} and \textbf{COCO-SP}.
We report the average and standard deviation of each result over 5 runs. 


\subsection{Results}
\begin{table}[htbp]
	\centering
	\caption{The average (\%) ± std over 5 runs on node classification tasks.
    Bold: best; Underline: second-best.
    }
    \resizebox{\linewidth}{!}{ 
	\begin{tabular}{c@{ }c@{ }c@{ }c@{ }c}
		\hline
		\multirow{2}{*}{Model} & \multicolumn{4}{c}{Dataset}                                 \\ 
        \cline{2-5} 
		& ogbn-arxiv   & ogbn-proteins & ogbn-products & pokec        \\ 
        & Accuracy $\uparrow$   & ROC-AUC $\uparrow$ & Accuracy $\uparrow$ & Accuracy $\uparrow$        \\ \hline
		GCN   & 75.20 ± 0.03 & 72.51 ± 0.35 & 83.90 ± 0.10 & 62.31 ± 1.13 \\
		GAT   & 73.83 ± 0.18 & 68.08 ± 0.55 & 82.63 ± 0.07 & 68.47 ± 1.42 \\
		SIGN  & 75.80 ± 0.08 & 71.24 ± 0.46 & 80.98 ± 0.31 & 68.01 ± 0.25 \\
		SGC   & 74.30 ± 0.11 & 70.31 ± 0.23 & 81.21 ± 0.12 & 52.03 ± 0.84 \\
        \midrule
		DIFFormer & \underline{77.32 ± 0.11} & 79.49 ± 0.44 & 85.21 ± 0.62 & 69.24 ± 0.76 \\
		SGFormer & 76.97 ± 0.02 & \underline{79.53 ± 0.38} & 89.09 ± 0.10 & 73.76 ± 0.24 \\
		Polynormer & 76.94 ± 0.10 & 78.97 ± 0.47 & \underline{90.17 ± 0.16} & \underline{77.27 ± 0.23} \\
        \midrule
		Ours  & \textbf{77.45 ± 0.08} & \textbf{81.23 ± 0.47} & \textbf{90.37 ± 0.05} & \textbf{79.11 ± 0.06} \\
		\bottomrule
	\end{tabular}%
    }
	\label{experimental_result_node}%
\end{table}%

\begin{table*}[htbp]
  \centering
  \caption{The average ± std over 5 runs on graph-level tasks.
  Bold: best; Underline: second-best.
  }
     \resizebox{0.85\textwidth}{!}{ 
     \begin{tabular}{cccccc}
    \toprule
    \multicolumn{1}{c}{\multirow{2}[4]{*}{Model}} & \multicolumn{5}{c}{Dataset} \\
\cmidrule{2-6}          & \multicolumn{1}{p{6.5em}}{ogbg-molhiv} & \multicolumn{1}{p{6.5em}}{Peptides-func} & \multicolumn{1}{p{6.5em}}{ Peptides-struct} & \multicolumn{1}{p{6.5em}}{ PascalVOC-SP} & COCO-SP \\
          & ROC-AUC $\uparrow$ & AP $\uparrow$ & MAE $\downarrow$ & F1-score $\uparrow$ & F1-score $\uparrow$ \\
    \midrule
    GCN   & 0.7599 ± 0.0119 & 0.5930 ± 0.0023 & 0.3496 ± 0.0013 & 0.1268 ± 0.0060 & 0.0841 ± 0.0010 \\
    GIN   & 0.7707 ± 0.0149 & 0.5498 ± 0.0079 & 0.3547 ± 0.0045 & 0.1265 ± 0.0076 & 0.1339 ± 0.0044 \\
    GatedGCN & -     & 0.5864 ± 0.0077 & 0.3420 ± 0.0013 & 0.2873 ± 0.0219 & 0.2641 ± 0.0045 \\
    \hline
    GraphGPS & 0.7880 ± 0.0101 & 0.6535 ± 0.0041 & 0.2500 ± 0.0005 & 0.3748 ± 0.0109 & 0.3412 ± 0.0044 \\
    Exphormer & -     & 0.6258 ± 0.0092 & 0.2512 ± 0.0025 & 0.3446 ± 0.0064 & 0.3430 ± 0.0108 \\
    GRIT  & -     & \underline{0.6988 ± 0.0082} & \underline{0.2460 ± 0.0012} & -     & - \\
    GECO  & \underline{0.7980 ± 0.0200} & 0.6975 ± 0.0025 & 0.2464 ± 0.0009 & \underline{0.4210 ± 0.0080} & 0.3320 ± 0.0032 \\
    \hline
    GPS+Mamba & -     & 0.6624±0.0079 & 0.2518±0.0012 & 0.4180±0.0120 & 0.3895±0.0125 \\
    GMN & -     & 0.6972±0.0100 & 0.2477±0.0019 & 0.4192±0.0108 & \underline{0.3909±0.0128} \\
    \hline
    Ours  & \textbf{0.8006 ± 0.0128} & \textbf{0.7131 ± 0.0057} & \textbf{0.2450 ± 0.0019} & \textbf{0.4375 ± 0.0029} & \textbf{0.3943 ± 0.0038} \\
    \bottomrule
    \end{tabular}%
    }
  \label{experimental_result_graph}%
\end{table*}%

Table \ref{experimental_result_node} and Table \ref{experimental_result_graph} present the experimental results of our model and all baselines. 
By leveraging the text features extracted by TAPE, all models exhibited commendable performance on \textbf{ogbn-arxiv}, as evidenced by a comparison with the official OGB leaderboard. 
Experimental results show that none of the GNN baselines achieve competitive performance, consistent with the well-known over-smoothing and over-squashing limitations of GNNs.
Existing Graph Transformer variants employing local-and-global or local-to-global attention schemes generally exhibit sub-optimal performance.
We also observe that the state space model GMN achieves sub-optimal performance on the COCO-SP dataset, probably because of its better capability in capturing long-range dependencies.
In contrast, G2LFormer consistently outperforms all baselines, proving the effectiveness of its global-to-local attention mechanism.

\subsection{Ablation study}
\begin{figure}[tbp]
	\centerline{\includegraphics[width=\linewidth]{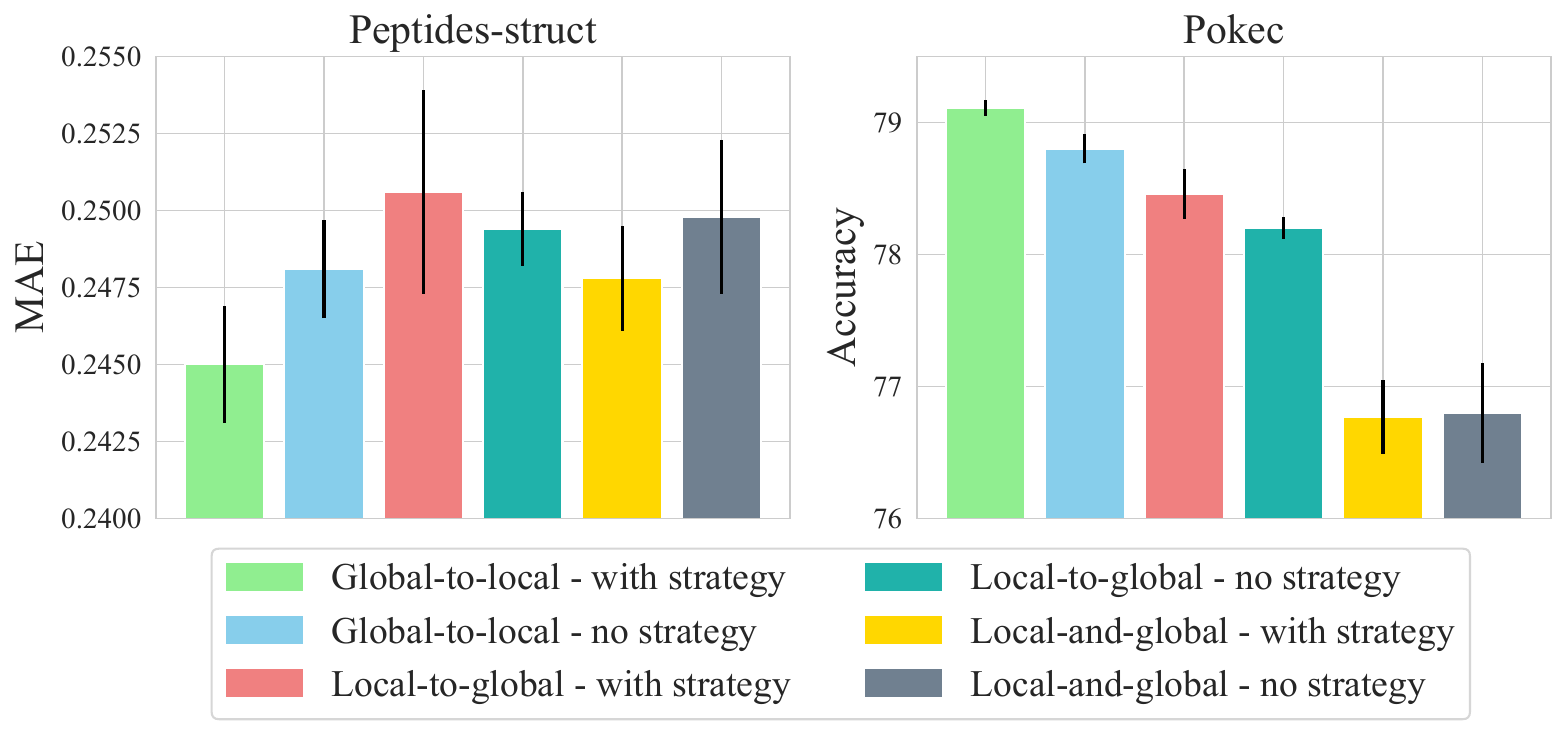}}
	\caption{
    Ablation study: 'with strategy' uses cross-layer information fusion strategy; 'no strategy' omits it.
    }
	\label{ablation_study}
\end{figure}

In this work, we employ the cross-layer information fusion strategy in the implementation of a global-to-local attention scheme. 
To systematically evaluate the contributions of each component, we conduct ablation studies examining (1) the attention scheme variants and (2) the cross-layer fusion strategy independently.
Although the local-and-global attention scheme is theoretically independent of cross-layer information fusion strategies, empirical evidence suggests that such strategies can enhance its effectiveness. 
Therefore, we empirically validate the influence of the cross-layer information fusion strategy on the local-and-global attention scheme, as well.
In summary, the experimental modules encompass three attention schemes: global-to-local scheme, local-to-global scheme, and local-and-global scheme, with or without the cross-layer information fusion strategy.
The performance of each module is evaluated on \textbf{Peptides-struct} and \textbf{pokec}, with results presented in Figure \ref{ablation_study}.

Figure \ref{ablation_study} indicates that the cross-layer information fusion strategy employed in this work benefits the global-to-local and local-to-global attention schemes, which implies that the sequential output methodology between global layers and local layers is susceptible to information loss and requires corresponding measures to prevent it. Since both local-to-global and local-and-global attention schemes are suboptimal and incorporate more misleading semantic information, cross-layer information fusion strategies may fail to yield consistent performance improvements. 
Meanwhile, the global-to-local scheme incorporating a cross-layer information fusion strategy consistently maintains optimal performance, which is likely attributable to the alleviation of the over-globalizing problem in global layers. 

Our findings also align with previous research suggesting that existing attention schemes exhibit inherent limitations \cite{ijcai2023p501}, and the challenges associated with GTs underscore the exploration of novel attention schemes \cite{pmlr-v235-xing24b}. 
Note that this work primarily focuses on demonstrating the feasibility of the global-to-local attention scheme. 
Its optimality remains to be fully investigated, as the expressiveness of attention mechanisms is typically backbone-model-dependent.

\subsection{Efficiency}

\begin{table}[htbp]
	\centering
	\caption{Efficiency comparison including training time and GPU memory usage. ``OOM'' represents out-of-memory on the GPU with 24GB of memory.}
	\resizebox{\linewidth}{!}{ 
    \begin{tabular}{@{}c@{}c@{ }c@{ }c@{ }c@{ }c@{ }c@{ }c@{}}
		\toprule
		&       & \multicolumn{3}{c}{ogbn-proteins} & \multicolumn{3}{c}{pokec} \\
		\cline{2-8}
		& Batch\_size & 10000 & 30000 & 50000 & 100000 & 200000 & 300000 \\
		\midrule
		\multirow{5}[2]{*}{\begin{tabular}{c} Train/\\epoch \\ (s) \end{tabular}} & GCN   & 4.2774  & 3.9357  & 3.4786  & 3.5628  & 2.9714  & 2.8122  \\
		& Cluster-GCN & 4.7519  & 4.1797  & 3.5374  & 5.0635  & 4.4248  & 4.1593  \\
		& SGFormer & 4.1850  & 2.8559  & 2.5604  & 3.7412  & 3.6755  & 3.2985  \\
		& Polynormer & 6.3792  & 4.9687  & OOM   & 5.1300  & 4.8020  & 4.6152  \\
		& G2LFormer & 6.5012  & 5.8185  & 5.4463  & 5.8678  & 5.6356  & 5.0592  \\
		\midrule
		\multirow{5}[2]{*}{\begin{tabular}{c} Memory \\ (MB) \end{tabular}} & GCN   & 2019  & 5893  & 11741 & 1115  & 3197  & 6994 \\
		& Cluster-GCN & 3923  & 8769  & 17911 & 1855  & 6175  & 9514 \\
		& SGFormer & 568   & 1515  & 3458  & 1663  & 3131  & 4713 \\
		& Polynormer & 6504  & 16813 & OOM   & 3745  & 12301 & 21345 \\
		& G2LFormer & 3567  & 13307 & 18201 & 3055  & 8099  & 15661 \\
		\bottomrule
	\end{tabular}%
    }
	\label{efficiency}%
\end{table}%

To evaluate the efficiency of G2LFormer, we follow the approach of FairGT \cite{fairgt2024luo}. Table \ref{efficiency} presents the per-epoch training time and GPU memory usage of G2LFormer compared with representative baselines on the ogbn-proteins and pokec datasets.
When considering the performance advantages demonstrated in the experimental results section, G2LFormer has achieved a favorable balance between expressivity and computational cost. 
Although it incurs higher training time than SGFormer and GCN variants due to the additional cross-layer information fusion strategy, it consistently consumes less GPU memory (18.4-45.2\% lower) than Polynormer. 
Notably, Polynormer runs out of memory (OOM) on ogbn-proteins when the batch size reaches 50,000, whereas G2LFormer remains stable.
Moreover, the G2LFormer backbone can be further optimized or replaced with more efficient models to accelerate training in future work.



\subsection{Scalability}

\begin{figure}[h!]
	\centerline{\includegraphics[width=\linewidth]{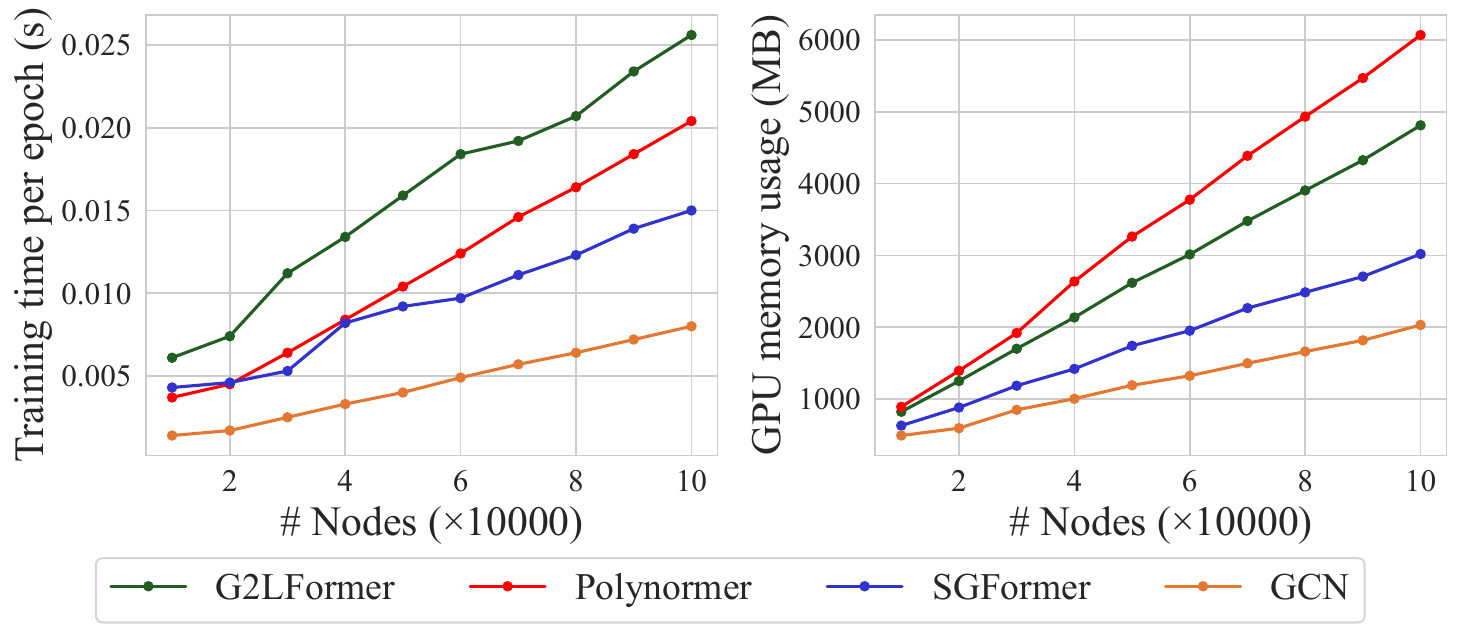}}
	\caption{Scalability test of training time and GPU memory usage.}
	\label{spacetime_results}
\end{figure}

To evaluate scalability, we test the proposed model on randomly generated Erdos-Renyi graphs with 10,000 to 100,000 nodes, each having an average degree of approximately 10 and 128-dimensional node features. 
Training runs for 1,000 epochs. 
Comparative scalability analysis includes Polynormer, SGFormer and GCN.
Figure \ref{spacetime_results} shows that G2LFormer's time and space complexity scale linearly with the number of nodes, confirming its $O(n)$ complexity.

\section{Conclusions}
Motivated by the limitations in existing attention schemes of GTs, this work introduces a global-to-local attention scheme. 
A cross-layer information fusion strategy is adopted, incurring an acceptable overhead to mitigate the problems of information loss and over-globalization in the integration of GTs and GNNs.
Building on the aforementioned insights, we propose the G2LFormer model. 
A variety of datasets are selected that encompass both node-level and graph-level tasks. 
Experimental results demonstrate that G2LFormer achieves state-of-the-art performance on most benchmark datasets while maintaining linear computational complexity. 
We aim to offer insights into the global-to-local attention scheme and encourage reconsideration of the trade-off between scalability and expressivity.

\bibliography{references}
\bibliographystyle{plain}

\end{document}